\documentclass[10pt,twocolumn,letterpaper]{article}

\usepackage{iccv}
\usepackage{times}
\usepackage{epsfig}
\usepackage{graphicx}
\usepackage{amsmath}
\usepackage{amssymb}

\usepackage{booktabs}
\usepackage{cases}
\usepackage{arydshln}

\usepackage[breaklinks=true,bookmarks=false,colorlinks]{hyperref}

\usepackage[capitalize]{cleveref}
\crefname{section}{Sec.}{Secs.}
\Crefname{section}{Section}{Sections}
\Crefname{table}{Table}{Tables}
\crefname{table}{Tab.}{Tabs.}
\usepackage{multirow}

\iccvfinalcopy 

\def\iccvPaperID{****} 

\ificcvfinal\fi

\begin{document}

\title{RecursiveDet: End-to-End Region-based Recursive Object Detection}

\author{$\text{Jing Zhao}^1$, $\text{Li Sun}^{1,2*}$, $\text{Qingli Li}^1$ \\
$^1$Shanghai Key Laboratory of Multidimensional Information Processing, \\
$^2$Key Laboratory of Advanced Theory and Application in Statistics and Data Science,\\ 
East China Normal University, Shanghai, China
}

\maketitle
\ificcvfinal\thispagestyle{empty}\fi

\begin{abstract}
   End-to-end region-based object detectors like Sparse R-CNN usually have multiple cascade bounding box decoding stages, which refine the current predictions according to their previous results. Model parameters within each stage are independent, evolving a huge cost. In this paper, we find the general setting of decoding stages is actually redundant. By simply sharing parameters and making a recursive decoder, the detector already obtains a significant improvement. The recursive decoder can be further enhanced by positional encoding (PE) of the proposal box, which makes it aware of the exact locations and sizes of input bounding boxes, thus becoming adaptive to proposals  from different stages during the recursion. Moreover, we also design  centerness-based PE to distinguish the RoI feature element and dynamic convolution kernels at different positions within the bounding box. To validate the effectiveness of the proposed method, we conduct intensive ablations and build the full model on three recent mainstream region-based detectors. The RecusiveDet is able to achieve obvious performance boosts with even fewer model parameters and slightly increased computation cost. Codes are available at \href{https://github.com/bravezzzzzz/RecursiveDet}{https://github.com/bravezzzzzz/RecursiveDet}.
\end{abstract}

\renewcommand{\thefootnote}{*} 
\footnotetext[1]{Corresponding author, email: sunli@ee.ecnu.edu.cn. This work is supported by the Science and Technology Commission of Shanghai Municipality under Grant No. 22511105800, 19511120800 and 22DZ2229004.}

\section{Introduction}
Object detection has been intensively investigated by computer vision community for decades. Traditional detectors built by deep convolutional neural network (CNN) are either anchor-based \cite{girshick2015fast,ren2015faster,liu2016ssd} or anchor-free \cite{redmon2016you,tian2019fcos,zhou2019objects}. The former performs classification and regression based on pre-defined densely tiled bounding boxes, while the latter only assumes 
grid points in the 2D image plane. On the other hand, detection can be completed in a single stage, two stages or even multiple cascade stages. The single-stage method directly gives predictions without further 
modifications, which is usually simple and efficient. Two- or multi-stage methods repeatedly make corrections based on previous results, which offer better results but cost more model parameters and calculations. Except for the first stage, later stages are usually region-based, focusing on the local region within the bounding box, which is often realized by RoI Align \cite{he2017mask}. 

\begin{figure}[t]
    \centering
    \includegraphics[width=0.7\columnwidth]{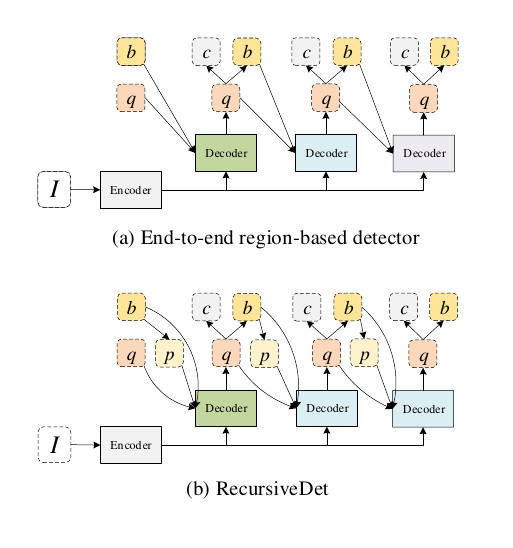}
    \caption{Comparison between end-to-end region-based detector and RecusiveDet.
    $I$ is the input image feature. $q$ and $b$ are proposal feature and proposal box at input. $c$ and $b$ indicate the class and bounding box of the predictions at output. In RecursiveDet, we encode $b$ into PE vector $p$.
    (a) Detectors like Sparse R-CNN, AdaMixer and DiffusionDet have cascade stages, which bring a huge amount of parameters. (b) Our proposed RecursiveDet simply shares the decoder parameters and makes the structure recursive, which reduces the model size.
    }
    \label{fig:intro}
\end{figure}

Although these detectors have been widely used, 
they are often complained for complex pre- and post-designs, \emph{e.g.}, anchor sizes, standards for positive and negative samples, and non-maximum suppression (NMS) on results. DEtection TRansformer (DETR) simplifies the framework. 
It 
relies on multi-stage transformer decoder 
layers to update learnable queries into object features, which can be 
decoded into bounding box predictions at each stage. During training, DETR utilizes bipartite matching to determine positive and negative query samples in a dynamic way. 
Since each ground truth (GT) box is assigned with only one query as positive, NMS is no longer necessary. However, DETR 
has extremely low efficiency, mainly due to the lack of inductive bias in the transformer decoder. Later works 
incorporate either bounding box positional encoding (PE), iterative refinement or local deformation into the decoder, hence greatly improve the training efficiency.

Similar to DETR, region-based methods \cite{sun2021sparse, gao2022adamixer, chen2022diffusiondet} also treat object detection as a set 
prediction task between predictions and GTs, and they all adopt a multi-stage strategy. Sparse R-CNN 
is a 
representative work depending on RoI Align. It employs a set of learnable proposal features to generate dynamic conv filters for image RoI 
features within 
corresponding proposal boxes. 
Compared to DETR, region-based methods obtain accurate results and high data efficiency. However, the model sizes are obviously larger than DETR-series, which not only limit their applications but also cause degradation for the small-scale dataset. 
We find that 
some specific layers in region-based methods account for most of the 
parameter number, so they can potentially 
be improved. Moreover, since proposal features are equivalent to object queries in DETR, it should be 
practical to introduce PE of bounding boxes like \cite{liu2022dab}, so the self attention and dynamic conv know the exact global and local positions of image features. 

Based on the above considerations, this paper proposes the RecursiveDet, a common cascade decoder structure built on dynamic conv for region-based detectors. Specifically, we share the decoder parameters and make the same module recursive in different stages, 
allowing training it
based on proposal 
boxes of different qualities. 
By simply sharing all 
decoder parameters among different stages, 
the model size is apparently reduced while its performance only suffers a slight drop. Moreover, since the dynamic layer for generating conv kernels and out layer after dynamic conv are huge, we intend to make full use of them 
and apply the dynamic conv for more layers.
In practice, we update the proposal feature after the first group of dynamic convs. Then the updated feature is given to the same dynamic layer, specifying conv kernels for the second group. 
In other words, the dynamic layer and out layer are reused to increase the model depth within a stage without extra parameter cost.  

Due to sharing parameters of decoder stages, 
its input 
becomes 
indispensable to distinguish each decoding stage. 
Therefore, more information about the previous predictions 
needs to be utilized as 
input. Inspired by \cite{meng2021conditional, wang2022anchor, liu2022dab}, we 
encode the positions and shapes of the bounding boxes into PE vectors, 
which are then employed 
for both 
position-aware self attention and dynamic conv. 
Since the PE of bounding box is only computed from the global image coordinate and shape size, 
we design a 
centerness-based PE to represent different locations within the RoI as 
compensation. By 
combining the PE of bounding box and centerness within dynamic conv, 
our RecursiveDet model is able to fully exploit both proposal features and boxes in the end-to-end region-based detector. 

To validate the effectiveness of the proposed RecursiveDet, intensive experiments and ablation studies are conducted on MS-COCO 
dataset. We implement 
RecursiveDet in three mainstream detectors, including Sparse R-CNN, AdaMixer and DiffusionDet, and achieve consistent performance boosts. We point that Sparse R-CNN and AdaMixer are end-to-end, while DiffusionDet is not. In summary, the contribution of this paper lies in the following aspects.
\begin{itemize}
    \item We propose to share the decoder parameters among different stages in the 
    region-based detector, which significantly reduces the model size without severely degrading the performance. Moreover, we repeatedly use 
    the dynamic layer to generate conv kernels for more layers in a single stage, increasing the model depth for further performance gain.
    \item We design the bounding box PE according to the geometry information from predictions. Besides, 
    the local coordinate is represented based on centerness PE, discriminating each feature within RoI. We make these PEs to 
    participate in self attention and dynamic conv 
    in the region-based detector. 
    \item 
    The proposed RecursiveDet is implemented in different end-to-end region-based detectors, showing that it enhances all of them 
    and reduces the model size at the same time.
\end{itemize}
\section{Related Work}
\noindent{\textbf{CNN-based detector}.}
Earlier CNN-based object detectors 
hold 
assumptions on densely located anchor boxes \cite{girshick2015fast,liu2016ssd,lin2017feature,cai2018cascade} or gird points \cite{law2018cornernet,zhu2019feature,tian2019fcos,yang2019reppoints} in 2D plane, 
and they compute the classification and bounding box regression loss for each candidate sample. 
Anchor-based methods heavily rely on hyper-parameters of anchor sizes. Moreover, they need heuristic rules, like the IoU threshold, to determine positive and negative anchors. The NMS post operation is also cumbersome. Grid point-based methods make detectors free from anchor boxes. They 
directly locate 
the GT box from the grid, however, positive and negative assignment is still a critical issue, which is often realized in a simple way. \emph{E.g.}, FCOS \cite{tian2019fcos} assumes the points inside a GT box to be positive, and RepPoints \cite{yang2019reppoints} labels a grid point nearest to the GT box as the positive. Giving the importance of sample assignment, researchers propose different schemes \cite{zhang2020bridging,kim2020probabilistic,ge2021ota} to prevent heuristic rules. Another way to characterize different detectors is based on the number of detection stages. Single stage detectors \cite{liu2016ssd,redmon2016you, redmon2017yolo9000} are more efficient than two- \cite{girshick2015fast,ren2015faster, dai2016r} or multi-stage \cite{cai2018cascade} competitors. They directly give results from whole image, while two- or multi-stage methods bring region-based 
feature from RoI Align to later stages to 
increase the accuracy.

\noindent{\textbf{Detection transformer}.} DETR \cite{carion2020end} opens a new era for object detection, since it eliminates most handcrafted designs. 
It starts from a set of learnable queries and employs multiple decoder attention blocks to update them into object features for the detection head. DETR adopts dynamic bipartite matching to build one-to-one relation between the GTs and predictions. However, the slow convergence and large training set requirement become its obstacle. Later works \cite{zhu2020deformable,gao2021fast,yao2021efficient,meng2021conditional,wang2022anchor,liu2022dab,li2022dn,zhang2022dino,chen2022recurrent} focus on incorporating spatial prior or extra query group into model for faster training and better results. Deformable DETR \cite{zhu2020deformable} utilizes deformable attention \cite{dai2017deformable, zhu2019deformable} and iterative refinement. SMCA \cite{gao2021fast} predicts a spatial mask from query to modulate attention matrix. REGO \cite{chen2022recurrent} employs RoI Align in DETR-series. 
Conditional \cite{meng2021conditional} and Anchor \cite{wang2022anchor} DETR encode the center of bounding box as PE to assist the decoder, and DAB-DETR \cite{liu2022dab} further extends PE to 
bounding box coordinates including width and height. DN-DETR \cite{li2022dn} introduces de-noising query group to accelerate training. DINO \cite{zhang2022dino} improves it through contrastive learning and query selection. 
 
\noindent{\textbf{End-to-end region-based detector}. }
Similar to DETR-series, end-to-end region-based detectors \cite{sun2021sparse,gao2022adamixer,zheng2022progressive,zhao2022iou,chen2022diffusiondet} start from learnable proposal features and update them into object features for final predictions in a cascade manner. They also borrow the dynamic bipartite matching in DETR to prevent NMS. However, instead of cross attention, dynamic conv is used to connect image with proposal features. Moreover, 
such detectors sample RoI feature within the bounding box from the previous stage for dynamic conv. The first representative work is Sparse R-CNN \cite{sun2021sparse}. To further speed up the convergence, 
AdaMixer \cite{gao2022adamixer} replaces RoI Align by sampling points directly from the 3D feature pyramid and borrows the idea of MLP-Mixer \cite{tolstikhin2021mlp} which changes axis of the second time dynamic conv. DiffusionDet \cite{chen2022diffusiondet} first utilizes training loss in diffusion model, and builds a multi-step multi-stage detector based on Sparse R-CNN. Our RecursiveDet belongs to this type, and it can be applied in most of them to improve the cascade decoding stages.

\noindent\textbf{Recursive module in deep model.} The typical recursive model in neural networks is RNN \cite{hochreiter1997long,chung2014empirical}, which is unrolled in multi-step during training. Many models \cite{kim2016deeply,liang2015recurrent,tai2017image} with recursive module have been proposed to address different tasks. Recent work \cite{shen2022sliced} applies recursive transformer block to increase the model depth. In object detection, recursive backbone \cite{liu2020cbnet} and FPN \cite{qiao2021detectors} have been proposed, however, none of the work deals with recursive 
decoder attention, particularly for object detection task.

\section{Method}
\subsection{Preliminaries}
This paper focuses on the end-to-end region-based detectors, among which Sparse R-CNN is the first typical model. It has a learnable proposal feature set $\mathcal{Q}=\{q_i|q_i\in\mathbb{R}^c\}$, in which each $q_i$ is a $c$ dimension vector to represent an object in image. Here $i=1,2,\cdots, N$ and $N$ is the total number of proposal features. Correspondingly, it keeps a set of proposal boxes $\mathcal{B}=\{b_i|b_i=(x,y,w,h)\}$ with the same number of $\mathcal{Q}$, 
where $b_i$ is a bounding box centered at $(x,y)$ with its width and height representing by $w$ and $h$. According to $\mathcal{B}$, the image RoI feature set $\mathcal{F}=\{f_i|f_i\in\mathbb{R}^{7\times 7\times c}\}$ is obtained from a multi-resolution feature pyramid by RoI Align operation, and each element is of the spatial size $7\times 7$. 
Sparse R-CNN has multiple cascade stages that gradually refine $\mathcal{B}$ to approach the GT box set. In each stage, the self attention within $\mathcal{Q}$ set is first computed. Then, the proposal feature $q_i$ is given to a huge dynamic layer,  providing a pair of dynamic conv kernels $k_i\in\mathbb{R}^{c\times d}$ and $v_i\in\mathbb{R}^{d\times c}$ for corresponding RoI 
feature $f_i$. 
Here $k_i $ and $v_i$ form a 
bottleneck,
which can be regarded as two successive layers consisting of $1\times 1$ conv 
whose parameters are specified as 
$k_i$ and $v_i$. $c$ and $d$ are the channel of input RoI feature and hidden dimension respectively, with $c\gg d$. After the dynamic convs, $f_i$ is updated into $f_i'$, which is then mapped into object feature $o_i$ through an out layer, 
so that it can be utilized 
by the detection head to give results. 

Sparse R-CNN performs the one-to-one dynamic interactions between $\mathcal{Q}$ and $\mathcal{B}$. In each stage, it not only predicts the 
bounding boxes, but also updates $\mathcal{Q}$ and $\mathcal{B}$. 
Particularly, 
object feature $o_i$ 
updates proposal feature $q_i$, preparing for the next stage, and bounding box 
prediction 
also updates 
$b_i$ as the next stage input. During training, Sparse R-CNN dynamically computes a cost matrix between the predictions and GTs. It carries out bipartite matching based on it, so that the GT box is only assigned with one prediction. 
The one-to-one sample assignment strategy prevents it from NMS post processing. The cost and training loss are the same and they can be formulated as $\mathcal{L}=\lambda_{cls}\mathcal{L}_{cls}+\lambda_{L_1}\mathcal{L}_1+\lambda_{giou}\mathcal{L}_{giou}$, where $\mathcal{L}_{cls}$ is the focal BCE loss \cite{lin2017focal} for classification, $\mathcal{L}_1$ and $\mathcal{L}_{giou}$ are L1 and generalized IoU loss \cite{rezatofighi2019generalized} for regression.

\begin{figure*}[ht]
    \centering
    \includegraphics[width=0.8\textwidth]{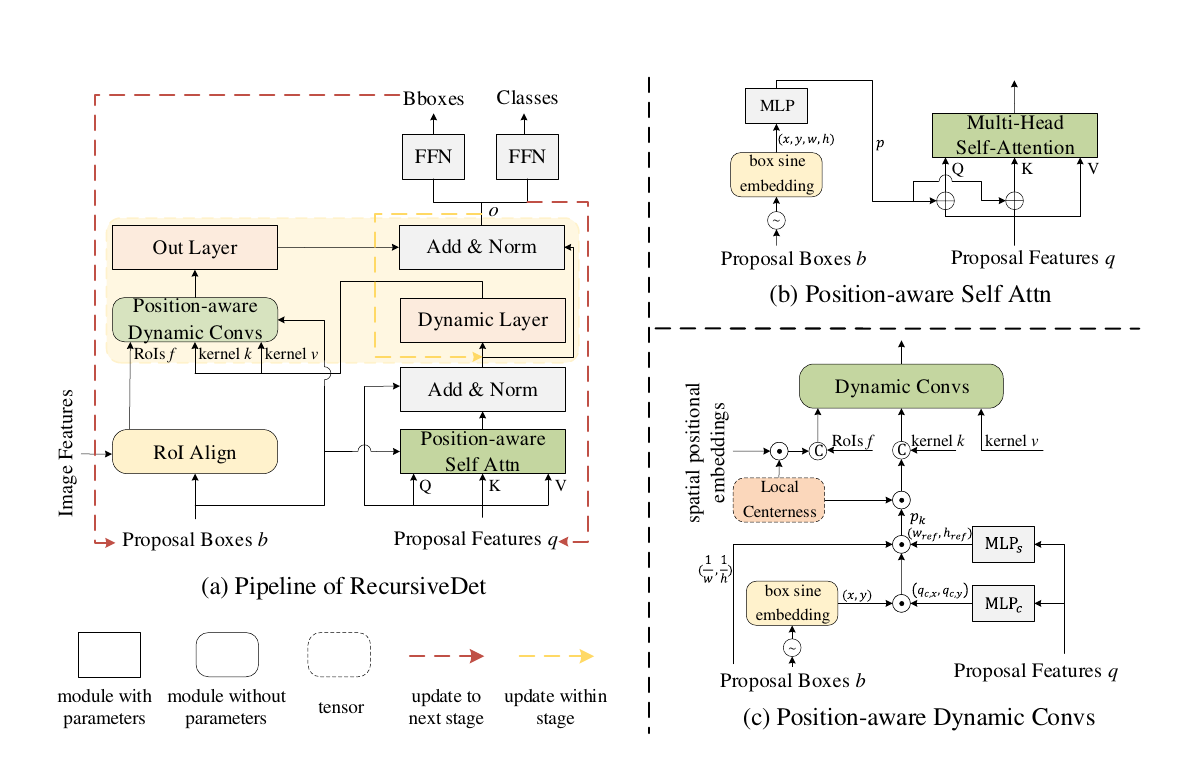}
    \caption{The details about decoder structure in 
    RecursiveDet. (a) An overview of our 
    method. We share the decoder parameters among 
    different stages and make it 
    recursive. The Dynamic layer and Out layer repeat 
    twice for increasing the depth of decoder. The object feature $o$ 
    and predicted bounding boxes are updated to the next stage as proposal features $q$ and proposal boxes $b$. The bounding boxes are encoded to 
    PE vectors $p$ and $p_k$ to participate in position-aware self attention and dynamic conv in (b) and (c), respectively. As shown in (c), the local coordinate 
    within RoI is calculated into a centerness mask to assist the bounding boxes PE in dynamic conv.
    }
    \label{fig:pipeline}
\end{figure*}

Besides Sparse R-CNN, AdaMixer and DiffusionDet are also region-based detectors,the former is end-to-end while the latter is not. They have similar cascade structures consisting of dynamic convs whose kernels are provided from dynamic layer based on proposal feature set $\mathcal{Q}$. Our RecursiveDet shares the basic computation pipeline with all of them, which is shown in \cref{fig:pipeline} (a). And it is compatible with all of them, with a 
significant 
reduction on the model size and increase on the result accuracy.

\subsection{Recursive Structure for Decoder}
Compared with DETR-series, end-to-end region-based detectors usually have a large decoder, which obviously expands their model size. The decoder $Dec^t$ at stage $t$, parameterized by $\theta^t$, takes the current proposal set $\mathcal{Q}^t$ and box set $\mathcal{B}^t$ as input, and updates them into $\mathcal{Q}^{t+1}$ and $\mathcal{B}^{t+1}$ for the next stage. We formalize the decoder in \cref{eq:eq3}. Note that RoI feature set $\mathcal{F}^t$ is also utilized by $Dec^t$.
\begin{equation}\label{eq:eq3}
    [\mathcal{Q}^{t+1},\mathcal{B}^{t+1}] = Dec^t(\mathcal{Q}^t,\mathcal{B}^t,\mathcal{F}^t;\theta^t)
\end{equation}

The key module in $Dec$ is the dynamic conv, and it builds the connection 
between proposal feature $q_i$ and image RoI feature $f_i$. However, we find 
one distinction between cross attention and dynamic conv. The former is usually light-weight, while the latter depends on a heavy dynamic layer $Dyn$, 
which 
maps 
$q_i$ into two sets of kernels $k_i$ and $v_i$, as is shown in \cref{eq:eq1}. 
\begin{equation}\label{eq:eq1}
    [k_i,v_i] = Dyn(q_i)=FC_{c\rightarrow 2c\times d }(q_i)
\end{equation}
Actually, $Dyn$ is a single FC layer with its input dimension $c$ 
the same as $q_i$, and output dimension equals $2c\times d$, which accommodates both $k_i$ and $v_i$. It accounts for a large percentage of 
model parameters 
in $Dec$. 
Another module, which 
costs many params and does not even exist in DETR-series, is the out layer $Out$. It is also an FC layer 
for reducing the spatial dimension from $7\times 7$ to $1$, as is illustrated in \cref{eq:eq2}. The input of $Out$ is the updated RoI feature $f_i'$ and the output is object feature $o_i$, which is utilized by the detection head for prediction and 
the next stage $Dec^{t+1}$ as its input proposal feature. 
\begin{equation}\label{eq:eq2}
    o_i = Out(f_i')+q_i=FC_{7\times 7\times c\rightarrow c }(f_i')+q_i\quad q_i^{t+1}=o_i
\end{equation}
The module details within $Dec$, particularly, the connection between different modules are illustrated in \cref{fig:pipeline} (a).

In following \cref{tab:params_of_decoder}, 
the number of parameters in a single stage of the decoder is listed. We can see that 
a single $Dyn$ has 8.4M parameters, therefore, its parameter cost in six different stages already surpasses the full model of 
DETR. 
The second largest module is $Out$, and it costs 3.2M parameters in Sparse R-CNN and DiffusionDet, and 8.4M in AdaMixer. Note that the multi-head cross or deformable attention layer in DETR-series is much smaller than $Dyn$. 

\begin{table}[ht]
\centering
\resizebox{\linewidth}{!}{
\begin{tabular}{@{}ccccccc@{}}
\toprule
Method & MSA & $Dyn$/MCA & $Out$ & Head & Others & Total\\ \midrule
 DETR & 0.3M & 0.3M & - & 26.0K$^*$ & 1.0M & 1.6M \\
 Deformable DETR & 0.3M & 0.2M & - & 0.2M & 0.5M & 1.2M  \\
 DAB-DETR & 0.4M & 0.4M & - & 26.0K$^*$ & 1.2M &  2.0M  \\
 \hdashline
 Sparse R-CNN & 0.3M & 8.4M & 3.2M & 0.3M & 1.0M & 13.2M\\
 AdaMixer & 0.3M & 8.4M & 8.4M & 0.1M & 1.2M & 18.4M \\
 DiffusionDet & 0.3M & 8.4M & 3.2M & 0.3M & 1.6M & 13.8M\\ \midrule
\end{tabular}}
\caption{Analysis of parameters 
in a single decoder layer.
'MSA', '$Dyn$', 'MCA', '$Out$' and 'Head' indicate multi-head self attention, dynamic layer, multi-head cross attention, output layer and detection head. 'Others' includes ffn, layer norm, etc. Note that, the detection head in DETR and DAB-DETR is shared among all stages, 
thus results with '$^*$' are divided by 
total stage number.}
\label{tab:params_of_decoder}
\end{table}
Based on the above analysis, 
we make two simple but important modifications to the decoder. The principle is to fully exploit it, particularly the $Dyn$ and $Out$ layers. 
Considering the large size of $Dec$ 
and its cascade structures in multiple stages, we 
share 
their params $\theta$ among stages to 
reduce the model size, thus 
$Dec$ becomes recursive with the same module repeatedly appearing in different stages. However, the recursion 
may result in unstable gradients during training, which is commonly encountered by RNN. Supposing the loss in stage $t$ is $\mathcal{L}^t$, 
its gradient 
$\frac{\partial \mathcal{L}^t}{\partial q^t}$ takes its effect into stage $(t-1)$ 
since $o^{t-1}=q^t$, hence the Jacobian matrix 
$\frac{\partial q^t} { \partial q^{t-1}}$ keeps multiplying until the first stage, causing the possible gradient vanishing or exploding. 
We argue that the total recursion number is not big.
Moreover, the detection loss is also computed in the intermediate stages, so the recursion of $Dec$ does not bring instabilities during training. In practice, considering the difficulty of the first decoding stage, detecting the object from the initial whole image or randomly located bounding box, 
we keep it out of the recursion loop with unique parameters to boost the performance. This causes 
a slightly increased model size.

Besides the recursion in different stages, we additionally make full use of $Dyn$ and $Out$ layers by making a short recursion from object feature $o_i$ to the input of the dynamic layer $Dyn$ within the same stage. In other words, 
$o_i$ is given to $Dyn$ layer again for generating new kernels $k_i'$ and $v_i'$. And they are used to perform dynamic conv 
on RoI $f_i$ again and update it into $f_i''$, which is then given to $Out$ to 
specify $o_i'$. We emphasize that the in-stage recursion does not increase the model size, since all trainable modules 
in the recursive loop, such as $Dyn$ and $Out$, share the same parameters. It only increases the amount of computation. 

\subsection{Bounding Box Positional Encoding}
By sharing 
$Dec$ 
among stages, the model gets compact and already achieves a better result. However, due to the common model parameters, $Dec$ treats input 
$q_i$ and 
$b_i$ from different stages in the same way, which could reduce its adaptability and be a limitation for the performance. 
We intend to introduce more information from the previous stage and make a compensation, 
so that $Dec$ has a sense of decoding 
stages and adjusts itself accordingly. 
Note that for DETR-series, it is common to give bounding box PEs to decoder. 
They help 
better model the relation between queries and keys 
during self and cross attention, therefore accurately representing objects. 
However, PE is usually ignored by end-to-end region-based detectors, mainly due to the destruction of global 
position by RoI Align. 

We argue that PE of the bounding box can still be valid even for region-based detectors. 
First, detectors like Sparse R-CNN still need self attention to model the pairwise similarities among $q_i \in \mathcal{Q}$, which can be effectively measured between the spatial positions of two bounding boxes. Second, 
the dynamic kernels 
can be more adaptive and directly related to location and shape. 
So we build the position-aware self attention and dynamic conv modules, as is shown in \cref{fig:pipeline} (b) and (c). 
We adopt a similar strategy for self attention with \cite{liu2022dab}. It first maps the 4D vector $(x,y,w,h)$ 
to 
sinusoidal embedding. Then it is further projected into a PE vector $p\in \mathbb{R}^c$ by an MLP, and $p$ is finally added onto the query and key tokens before self attention. For dynamic conv, the center 
$(x,y)$ and box shape $(w,h)$ are encoded in a separate manner. Utilizing two MLPs, denoted by $\text{MLP}_c$ and $\text{MLP}_s$, proposal feature $q$ is mapped into two vectors $q_c$ and $q_s$ in the geometry space, reflecting 
the center and shape of bounding box, respectively. The computations of $q_c$ and $q_s$ are provided in \cref{eq:eq5}. Here $q_s$, consisting of two scalars $w_{ref}$ and $h_{ref}$, is predicted from $q$ by $\text{MLP}_s$. $q_{c,x}$ and $q_{c,y}$ together form $q_c$, and each contributes half of the full dimension.
\begin{equation}\label{eq:eq5}
    \begin{aligned}
        q_c=[q_{c,x},q_{c,y}]=\text{MLP}_c(q)\in \mathbb{R}^c\\
        q_s=[w_{ref},h_{ref}]=\text{MLP}_s(q)\in \mathbb{R}^2
    \end{aligned}
\end{equation}
Then, they are modulated by geometry features from $(x,y)$ and $(w,h)$, specifying $p_{k}$ for kernels. We formalize this in \cref{eq:eq4}. Here $\odot$ means the Hadamard product. $p_{k}(x,w)$ and $p_k(y,h)$ are finally concatenated into $p_k$. 
\begin{equation}\label{eq:eq4}
\begin{aligned}
    p_{k}(x,w) = \frac{w_{ref}}{w}\text{Sinusoidal}(x)\odot q_{c,x}\\ 
    p_{k}(y,h) =  \frac{h_{ref}}{h}\text{Sinusoidal}(y)\odot q_{c,y}
\end{aligned}
\end{equation}
At the same time, image feature at each coordinate is also assigned with a PE vector according to the Sinusoidal function. As a result, each element in RoI feature $f_i$ has its unique positional representation, denoted by $p_{f}$. Note that $p_{f}$ and $p_{k}$ have the same dimensions, and they are concatenated with $f$ and $k$, respectively, completing the position-aware dynamic conv. 

\begin{table*}[htbp]
\centering
\begin{tabular}{@{}c|ccc|cccccc@{}}
\toprule
Method & Backbone & Epoch & Params & AP & $\rm{AP}_{50}$ &  $\rm{AP}_{75}$ & $\rm{AP}_{S}$ & $\rm{AP}_{M}$ & $\rm{AP}_{L}$\\ \midrule
DETR \cite{carion2020end} & ResNet-50 & 500 & 41M & 42.0 & 62.4 & 44.2 & 20.5 & 45.8 & 61.1\\
Deformable DETR \cite{zhu2020deformable} & ResNet-50 & 50 & 40M & 43.8 & 62.6 & 47.7 & 26.4 & 47.1 & 58.0\\
Conditional DETR \cite{meng2021conditional} & ResNet-50 & 50 & 44M & 40.9 & 61.8 & 43.3 & 20.8 & 44.6 & 59.2 \\
DAB-DETR \cite{liu2022dab} & ResNet-50 & 50 & 44M & 42.2 & 63.1 & 44.7 & 21.5 & 45.7 & 60.3 \\
 DN-DETR \cite{li2022dn} & ResNet-50 & 50 & 44M & 44.1 & 64.4 & 46.7 & 22.9 & 48.0 & 63.4 \\
\hdashline
Sparse R-CNN \cite{sun2021sparse} & ResNet-50 & 36 & 106M & 45.0 & 63.4 & 48.2 & 26.9 & 47.2 & 59.5 \\
AdaMixer \cite{gao2022adamixer} & ResNet-50 & 36 & 135M & 47.0 & 66.0 & 51.1 & 30.1 & 50.2 & 61.8 \\
DiffusionDet \cite{chen2022diffusiondet} & ResNet-50 & 60 & 111M & 45.8 & 65.6 & 49.2 & 29.7 & 48.6 & 61.1 \\
 RecursiveDet (Sparse R-CNN) & ResNet-50 & 36 & 55M & 46.5 & 65.4 & 50.9 & 29.7 & 49.1 & 59.9 \\
 RecursiveDet (AdaMixer) & ResNet-50 & 36 & 43M & 47.9 & 66.7 & 52.2 & 32.5 & 50.7 & 61.9 \\
 RecursiveDet (DiffusionDet) & ResNet-50 & 60 & 57M & 47.1 & 66.8 & 51.0 & 30.9 & 49.3 & 62.3 \\
 \midrule
 DETR \cite{carion2020end} & ResNet-101 & 500 & 60M & 43.5 & 63.8 & 46.4 & 21.9 & 48.0 & 61.8\\
Deformable DETR\ddag \cite{zhu2020deformable} & ResNet-101 & 50 & 59M & 47.2 & 66.6 & 51.1 & 28.5 & 50.9 & 62.4\\
Conditional DETR \cite{meng2021conditional} & ResNet-101 & 50 & 63M & 42.8 & 63.7 & 46.0 & 21.7 & 46.6 & 60.9 \\
DAB-DETR \cite{liu2022dab} & ResNet-101 & 50 & 63M & 43.5 & 63.9 & 46.6 & 23.6 & 47.3 & 61.5 \\
 DN-DETR \cite{li2022dn} & ResNet-101 & 50 & 63M & 45.2 & 65.5 & 48.3& 24.1 & 49.1 & 65.1 \\ \hdashline
Sparse R-CNN \cite{sun2021sparse} & ResNet-101 & 36 & 125M & 46.4 & 64.6 & 49.5 & 28.3 & 48.3 & 61.6 \\
AdaMixer \cite{gao2022adamixer} & ResNet-101 & 36 & 154M & 48.0 & 67.0 & 52.4 & 30.0 & 51.2 & 63.7 \\
DiffusionDet \cite{chen2022diffusiondet} & ResNet-101 & 60 & 130M & 46.6 & 66.3 & 50.0 & 0.0 & 49.3 & 62.8 \\
 RecursiveDet (Sparse R-CNN) & ResNet-101 & 36 & 74M & 47.1 & 65.7 & 51.8 & 29.3 & 50.6 & 61.5 \\
 RecursiveDet (AdaMixer) & ResNet-101 & 36 & 62M & 48.9 & 67.8 & 53.1 & 32.1 & 52.2 & 63.8 \\
 RecursiveDet (DiffusionDet) & ResNet-101 & 60 & 76M & 46.9 & 66.7 & 50.3 & 29.5 & 49.8 & 62.7  \\
 \midrule
 Sparse R-CNN \cite{sun2021sparse} & Swin-T & 36 & 110M & 47.9 & 67.3 & 52.3 & - & - & - \\
 AdaMixer \cite{gao2022adamixer} & Swin-S & 36 & 160M &51.3 & 71.2 & 55.7 & 34.2 & 54.6 & 67.3 \\
 
 Sparse R-CNN \cite{sun2021sparse} & Swin-B & 36 & 169M & 52.0 & 72.2 & 57.0 & 35.8 & 55.1 & 68.2 \\
 RecursiveDet (Sparse R-CNN) & Swin-T & 36 & 58M & 48.9 & 68.2 & 53.6 & 33.1 & 51.3 & 63.2 \\
 RecursiveDet (AdaMixer) & Swin-S & 36 & 68M & 52.1 & 71.6 & 56.9 & 36.2 & 55.1 & 68.3 \\
 
 RecursiveDet (Sparse R-CNN) & Swin-B & 36 & 118M & 53.1 & 73.0 & 58.4 & 37.0 & 56.6 & 69.5 \\
\bottomrule
\end{tabular}
\vspace{0.2cm}
\caption{Main results and comparisons with other object detectors on COCO 2017 val set. DETR uses 100 object queries, DiffusionDet and RecusiveDet built on it use 500 proposal features, and all other detectors use 300 of that. The results are from the original paper, mmdetection\cite{chen2019mmdetection} and Detectron2 \cite{Detectron2018}.
"\ddag" is the enhanced version with iterative box refinement and two-stage processing. } 
\label{tab:main_results}
\end{table*}

\subsection{Centerness-based Positional Encoding}\label{sec:centerness}
The bounding box PE for self attention and dynamic conv described in the previous section encodes $(x,y,w,h)$ in the global image coordinate. Thus, for each RoI feature $f$, the generated 
kernels $k$ and $v$ are shared by all elements within it. 
We intend to further enhance the adaptability of dynamic conv by expanding kernels $k$ to $k_e\in\mathbb{R}^{7\times 7\times c\times d}$, therefore, 
equipping each 
element in $f$ 
with unique conv kernels. 
Based on this motivation, we propose a centerness-based PE, 
which encodes the local coordinate in bounding box $b$ for 
feature $f$ and 
kernels $k$. Particularly, 
a single channel centerness \cite{tian2019fcos} mask 
$m\in\mathbb{R}^{7\times 7}$ defined by \cref{eq:eq6} is first calculated. Here $(x^*,y^*)$ denotes the local coordinate within the bounding box, and $x^*,y^*=0,1,\cdots,6$ have the same value range. 
\begin{equation}\label{eq:eq6}
    m(x^*,y^*)=\sqrt{\frac{\min(x^*,6-x^*)}{\max(x^*,6-x^*)} \frac{\min(y^*,6-y^*)}{\max(y^*,6-y^*)}}
\end{equation}
Note that $m$ is between 0 and 1, 
and a higher value means $(x^*,y^*)$ is near the center of the bounding box. 
Then, we multiply $m$ onto the RoI feature $f$ and kernel $k$. Since $m$ has only one channel, it is replicated on channel dimension $c$ to match $f$. For $k$, $m$ needs replication on both channel $c$ and hidden dimension $d$. $k$ also needs expansions on spatial dimension before the multiplication with $m$, enlarging its size to $7\times 7$. After the modulation by $m$, dynamic kernel $k$ is adapted into $k_e$, giving each element in $f$ a unique kernel. Our centerness-based PE is also shown in \cref{fig:pipeline} (c). 

The centerness mask $m$ is static with the same size of RoI feature $f$,
which means it has the same value for different $f_i$ and $k_i$, and it always gives the largest value at the center position. In practice, we try following two strategies to enhance it. First, we make it learnable as model parameters, with its initialization in \cref{eq:eq6}. Second, we adjust the largest value position by predicting the center coordinate from proposal feature $q$. However, we find the static $m$ strategy gives better results than the other two. 
Details are provided in the ablation study in \cref{sec:centerness_form}.


\begin{table*}[htbp]
\centering
\begin{tabular}{@{}c|cc|cccccc@{}}
\toprule
Method & Backbone & TTA & AP & $\rm{AP}_{50}$ &  $\rm{AP}_{75}$ & $\rm{AP}_{S}$ & $\rm{AP}_{M}$ & $\rm{AP}_{L}$\\ \midrule
AdaMixer \cite{gao2022adamixer} & ResNeXt-101-DCN &  & 49.8 & 69.3 & 54.3 & 30.0 & 52.1 & 64.3 \\
 Dynamic DETR \cite{dai2021dynamic} & ResNeXt-101-DCN &  & 49.3 & 68.4 & 53.6 & 30.3 & 51.6 & 62.5 \\
Sparse R-CNN & Swin-T & & 47.0 & 66.8 & 51.2 & 28.3 & 49.0 & 60.6\\
Sparse R-CNN & Swin-B &  & 52.2 & 72.6 & 57.2 & 32.3 & 54.9 & 67.2\\
 \midrule
RecursiveDet & Swin-T &  & 49.1 & 68.5 & 53.9 & 30.5 & 51.2 & 61.9 \\
 RecursiveDet & Swin-B &  & 53.1 & 73.9 & 58.5 & 33.7 & 55.8 & 67.9 \\
 RecursiveDet & Swin-B & \checkmark & 55.1 & 75.0 & 61.8 & 37.7 & 57.3 & 68.5 \\
\bottomrule
\end{tabular}
\vspace{0.2cm}
\caption{Results on COCO 2017 test-dev set. "TTA" indicates test-time augmentations, following \cite{zhang2020bridging}.}
\label{tab:results_test_dev}
\end{table*}

\section{Experiments}
\subsection{Implementation Details}
\noindent\textbf{Datasets.}\quad
We conduct extensive experiments on MS-COCO 2017 detection dataset \cite{lin2014microsoft} with 80 categories in total. There are about 118k images in the train2017 set and 5k 
in val2017. We report the standard MS COCO AP as the main evaluation metric.

\noindent\textbf{Training settings.}\quad
We adopt the same data augmentation as Sparse R-CNN, including random horizontal flipping, random crop and scale jitter of resizing the input images such that the shortest side is at least 480 and at most 800 pixels while the longest is at most 1333.
The training loss is the same as matching cost with loss weight 
$\lambda_{cls} = 2$, $\lambda_{L_1} = 5$ and $\lambda_{giou} = 2$. The learning rate is divided by 10 at 210K and 250K iterations if the  training schedule is 270k iterations, or at 350K and 420K if the training schedule is 450K. The default number of decoding stages is 
6.

\subsection{Main Results}
The proposed RecursiveDet is built on three region-based detectors, including Sparse R-CNN, AdaMixer and DiffusionDet. We provide the performance comparisons of RecursiveDet with well-established methods on COCO 2017 validation set in \cref{tab:main_results}. Note that, the sampling points in 3D feature space are out of the proposal boxes in AdaMixer, so the centerness-based PE is not implemented into it. It can be seen that the three region-based detectors exceed Cascade R-CNN and the other DETR-series detector, but require more model parameters. Owing to the recursive structure, our method saves a massive amount of parameters. With ResNet-50 backbone, RecursiveDet built on Sparse R-CNN 
outperforms Sparse R-CNN by 1.5 AP, while parameters are reduced from 106M to 55M. The models built on DiffusionDet and AdaMixer beat their baseline 1.6 AP (47.1 vs. 45.5) and 0.9 AP (47.0 vs. 47.9). 
RecursiveDet also behaves well when backbone scales up.  With ResNet-101, RecursiveDet achieves 47.1, 46.9 and 48.9 based on Sparse R-CNN, DiffusionDet and Adamixer. 
More surprisingly, RecursiveDet(AdaMixer) and RecursiveDet(Sparse R-CNN) reach 52.1 and 53.1 with the help of Swin-S and Swin-B, respectively. Although RecursiveDet(DiffusionDet) doesn't exceed its baseline, it reduces more than 30\% of the model size.
Note that, all our models save a large number of parameters. 
The results of our method on COCO test-dev set are listed in \cref{tab:results_test_dev}. RecursiveDet reaches 53.1 AP on Swin-B without bells and whistles. When test-time augmentation is utilized, it achives 55.1 AP.

\subsection{Ablation Study}
In this section, we use ResNet-50 
to perform ablation studies based on Sparse R-CNN architecture as default. The number of proposal features is 100.

\noindent\textbf{Recursive Structure.}\quad
Since Sparse R-CNN has multiple cascade decoder stages to refine the predictions progressively, the parameters of it are redundant. We propose to share them in different stages.
As shown in \cref{tab:ablation_recursive}, simply sharing the parameters and making the structure recursive reduces the model size from 106M to 40.1M, with a slight performance drop of 0.4 AP.
To take full advantage of the decoder's capabilities, 
we reuse the dynamic layer and out layer to increase the depth of model without adding new parameters. 
It gains 43.6 AP, 
surpassing the original Sparse R-CNN 0.8 AP. 
As the first stage is difficult to locate objects, 
we keep the initial stage independent with the other recursive ones.
Note that the total number of stages remains 6. The performance reaches 43.9 AP, gains 0.3 AP from model 'RecSt + RecDy' with a small cost and negligible latency.

\begin{table}[htbp]
\centering
\resizebox{\linewidth}{!}{
\begin{tabular}{@{}cccccccc@{}}
\toprule
RecSt & RecSt$^{\dag}$ & RecDy  &  AP & Params(M) & Flops(G) & L(ms) & FPS\\ \midrule
 &  &  &  42.8 & 106 & 134 & 49 & 20\\
 \checkmark &  &  &  42.4 & 40 & 134 & 48 & 21\\
 \checkmark &  & \checkmark &  43.6 & 40 & 142 & 50 & 20\\
  & \checkmark & \checkmark & 43.9 & 53 & 140 & 51 & 20\\ \bottomrule
\end{tabular}}
\vspace{0.2cm}
\caption{Ablation about recursive structure. 'RecSt' indicates that the parameters of different stages are shared, 'RecDy' means 
dynamic layer and out layer are reused. '$^{\dag}$' implies the first stage's parameters are independent of the remaining stages. }
\label{tab:ablation_recursive}
\end{table}

\noindent\textbf{Centerness form in PE.}\quad \label{sec:centerness_form}
As discribed in \cref{sec:centerness}, the local coordinate PE of centerness makes model explore location information more precisely. In addition to the static centerness computed within $7 \times 7$ region, it can be trained as learnable parameters of the model. The result in \cref{tab:ablation_centerness} shows a performance drop of about 0.3 AP. As different objects have various shapes and focus on different points, we predict a group of offsets to distinguish where the 
object center is. Let the value of centerness at this position be the maximum, the further away, the smaller. However, the result is unsatisfactory with 44.5 AP, which is 0.7 less than the static one.
\begin{table}[htbp]
\centering
\begin{tabular}{@{}cccc@{}}
\toprule
centerness & AP & $\rm{AP}_{50}$ &  $\rm{AP}_{75}$ \\ \midrule
static & 45.2 & 63.9 & 49.4 \\
learnable & 44.9 & 63.5 & 49.0 \\
 adjust & 44.5 & 63.3 & 48.4\\ \bottomrule
\end{tabular}
\vspace{0.2cm}
\caption{Variations of centernss-based PE. 'static' means the centerness is the same for different proposal boxes within the $7 \times 7$ region. 'learnable' makes the static centerness be learnable parameters. 'adjust' indicates 
a set of offsets is generated from RoI feature to adjust the position of the maximum value.}
\label{tab:ablation_centerness}
\end{table}

\noindent\textbf{Influences of different modules in RecursiveDet.}\quad
In this part, we 
analyze the effectiveness of different components in our model.
As shown in \cref{tab:ablation_components}, all modules proposed in this paper substantially contribute to the final results. The recursive structure improves the Sparse R-CNN from 42.8 AP to 43.9 AP. 
PE has not been 
introduced into region-based detectors before. 
We 
encode bounding boxes into embedding to participate in self attention and dynamic conv.
It achieves 0.3 AP performance gain. When different stages are shared, it is challenging for the decoder input to discriminate which stage is. Bounding box 
encoding plays its role by 
providing geometry information for decoder to distinguish the stage, and it gains 44.4 AP, exceeding the model of recursive structure by 0.5 AP.
And the time cost mainly comes from it.
Since the bounding box only has the global coordinates and shape sizes, we further employ centerness within the proposal box to introduce a local prior. With the help of centerness, the result finally reaches 45.2 AP.  

\begin{table}[htbp]
\centering
\resizebox{\linewidth}{!}{
\begin{tabular}{@{}cccccccc@{}}
\toprule
Recursive & bb-PE & cb-PE & AP & Params(M) &  Flops(G) & L(ms) & FPS\\ \midrule
 &  &  & 42.8 & 106 & 134 & 49 & 20\\
 \checkmark &  &  & 43.9 & 53 & 140 & 51 & 20 \\
  & \checkmark &  & 43.1 & 110 & 137 & 69 & 15\\
   & \checkmark & \checkmark & 43.4 & 110 & 138 & 72 & 14 \\
 \checkmark & \checkmark &  & 44.4 & 55 & 145 & 73 & 14\\
 \checkmark & \checkmark & \checkmark & 45.2 & 55 & 149 & 76 & 13\\\bottomrule
\end{tabular}}
\vspace{0.2cm}
\caption{Ablations about different components (Recursive structure, bounding-box PE(bb-PE) and centerness-based PE(Cb-PE)) in the proposed RecursiveDet.}
\label{tab:ablation_components}
\end{table}

\begin{figure}[htbp]
    \centering
    \includegraphics[width=0.8\columnwidth]{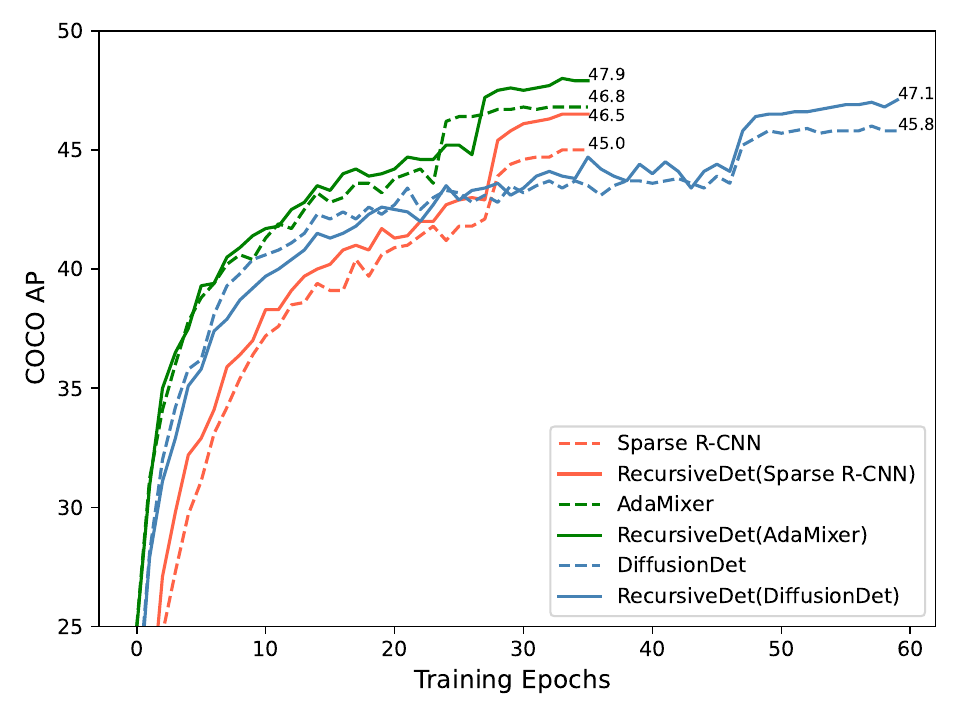}
    \caption{Convergence curves of Sparse R-CNN, AdaMixer, DiffusionDet and their counterpart in 
    RecursiveDet. All models are trained with  ResNet-50. Number of proposal feature is 300 for Sparse R-CNN and AdaMixer, and 500 for DiffusionDet.
    }
    \label{fig:convergence}
\end{figure}

\begin{figure}[htbp]
    \centering
    \includegraphics[width=0.8\columnwidth]{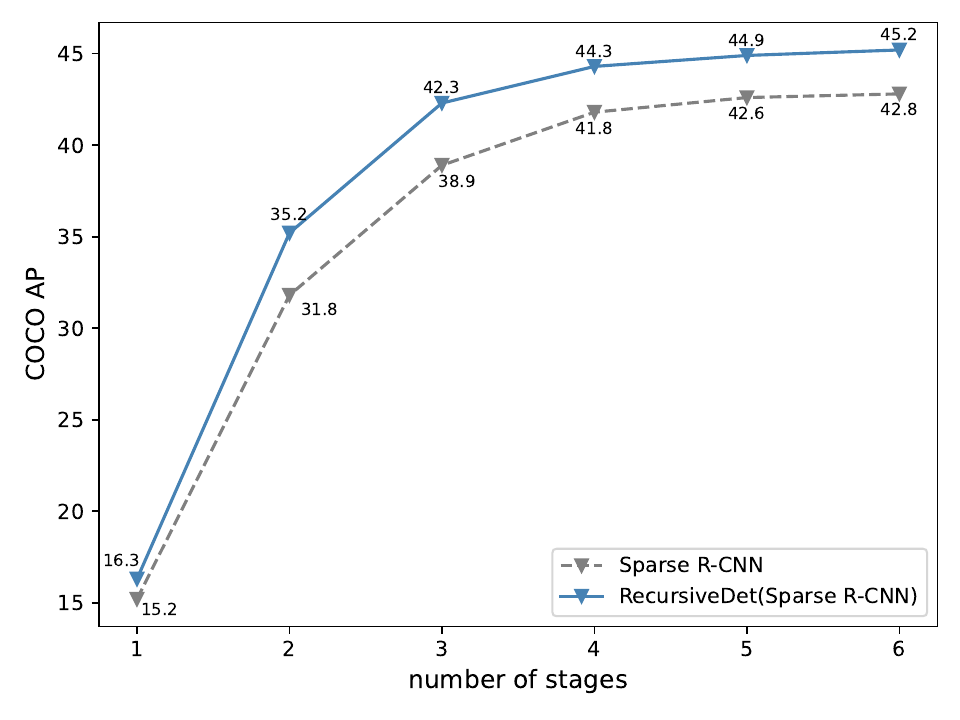}
    \caption{Effect of the number of stages in cascade and recursive structures. Both models are trained with six decoder stages.}
    \label{fig:stages}
\end{figure}

\noindent\textbf{Visualization.}\quad
In \cref{fig:convergence} we present the convergence speed of Sparse R-CNN, AdaMixer, DiffusionDet and RecursiveDet. It shows that our models are more efficient than their baselines. 
Since our method shares the decoder parameters, making the cascade detector recursive,
we compare the effectiveness of the number of decoder stages at inference between Sparse R-CNN and RecursiveDet in \cref{fig:stages}. It can be seen that our method outperforms Sparse R-CNN with any number of stages.

\section{Conclusion}
This paper investigates the 
region-based object detector. We propose a RecursiveDet, which increases the detection performance and reduces the model size. There are two types of recursion loops in our detector. 
We first share the decoder parameters and make it recursive among different stages. We also reuse the dynamic layer and out layer in decoder, and make a short in-stage recursion loop to increase the depth of model. To enhance the adaptability of the decoder, we design bounding box and centerness-based positional encoding, and further utilize them in self attention and dynamic conv. The RecursiveDet is implemented under three typical region-based detectors, including Sparse R-CNN, AdaMixer and DiffusionDet. It achieves consistent improvements on all of them with a lower cost on model parameters.

{\small
\bibliographystyle{ieee_fullname}
\bibliography{egbib}
}

\end{document}